\pdfoutput=1

\documentclass[11pt]{article}

\usepackage{acl}

\usepackage{times}
\usepackage{latexsym}

\usepackage[T1]{fontenc}

\usepackage[utf8]{inputenc}

\usepackage{microtype}

\usepackage{graphicx}
\usepackage{amssymb}
\usepackage{amsmath}
\usepackage{amsthm}
\usepackage{subfigure}
\usepackage{color}
\usepackage{algorithm}
\usepackage{algorithmic}
\usepackage{multirow}
\usepackage{bm}
\usepackage{bbm}
\newtheorem{myDef}{Definition}
\usepackage{bbding}
\usepackage{booktabs}
\usepackage{tabu}
\usepackage{microtype}
\usepackage{subfigure}
\usepackage{makecell}
\title{A Unified Temporal Knowledge Graph Reasoning Model Towards Interpolation and Extrapolation}

\author{
Kai Chen$^{1*}$,
Ye Wang$^{1}$\thanks{\, \, Equal contributions.} \ ,
Yitong Li$^{2 \dag}$,
Aiping Li$^1$\thanks{\, \, Corresponding authors.} \ ,
Han Yu$^1$,
Xin Song$^1$
\\
$^1$National University of Defense Technology, Changsha, China\\
$^2$Huawei Technologies Co., Ltd.\\
{\small \texttt{\{chenkai\_,ye.wang,liaiping,yuhan17,songxin\}@nudt.edu.cn} } \ {\small \texttt{liyitong3@huawei.com}}
}
\begin{document}
\maketitle
\begin{abstract}
Temporal knowledge graph (TKG) reasoning has two settings: interpolation reasoning and extrapolation reasoning. Both of them draw plenty of research interest and have great significance. 
Methods of the former de-emphasize the temporal correlations among facts sequences, while methods of the latter require strict chronological order of knowledge and ignore inferring clues provided by missing facts of the past.
These limit the practicability of TKG applications as almost all of the existing TKG reasoning methods are designed specifically to address either one setting. 
To this end, this paper proposes an original Temporal PAth-based Reasoning (TPAR) model for both the interpolation and extrapolation reasoning. 
TPAR performs a neural-driven symbolic reasoning fashion that is robust to ambiguous and noisy temporal data and with fine interpretability as well. Comprehensive experiments show that TPAR outperforms SOTA methods on the link prediction task for both the interpolation and the extrapolation settings. A novel pipeline experimental setting is designed to evaluate the performances of SOTA combinations and the proposed TPAR towards interpolation and extrapolation reasoning. More diverse experiments are conducted to show the robustness and interpretability of TPAR.

\end{abstract}

\section{Introduction}
Knowledge graph (KG) is a semantic network that represents real-world facts in a structured way using entities and relations \cite{Bollacker2008FreebaseAC,SongLTCL21,ZhaoJLJCW21}.
Typically, a fact is represented by a triple $(s,r,o)$ in KG, consisting of a subject entity $s$, an object entity $o$, and a relation $r$ between $s$ and $o$.

In the real world, many facts are closely associated with a particular time interval.
For example, the fact \emph{``Barack Obama is the president of USA''} is valid for the time period of \emph{2009 January 20th - 2017 January 20th } and the fact \emph{``Donald Trump is the president of USA''} is only valid for the following four years.
To represent such time-sensitive facts, Temporal Knowledge Graphs (TKGs) have recently gained significant attention from both academic and industrial communities. Specifically, TKGs extend static KGs by incorporating the temporal information $t$ into fact triples, represented as a quadruple $(s,r,o,t)$, which allows for modelling the temporal dependencies and evolution of knowledge over time, being crucial for reasoning time-evolving facts in applications such as financial forecasting, social networks, and healthcare.

TKG reasoning infers new knowledge with time-sensitive facts in existing TKGs, which generally has two settings: the interpolation reasoning \cite{xu2020tero,LacroixOU20,XuCNL21,ChenWLL22} and the extrapolation reasoning \cite{jin2020recurrent,SunZMH021,LiuMHJT22}.
Given a temporal knowledge graph with facts from time $t_0$ to time $t_T$, the interpolation reasoning infers missing facts at any time in history ($t_0 \leq t \leq t_T$) and the extrapolation reasoning attempt to predict unknown facts that may occur in the future ($t>t_T$).

Many approaches have been proposed to tackle the TKG reasoning problem, however, these two reasoning tasks are tackled in totally different and incompatible manners.
On the one hand, interpolation methods de-emphasize the temporal correlations among fact sequences while training,
thus it's difficult to cope with the challenges of invisible timestamps and invisible entities in extrapolation due to their poor inductive reasoning ability \cite{SunZMH021}.
On the other hand, most state-of-the-art (SOTA) extrapolation solutions require a strict chronological order of data during training. As a result, they can only predict unknown future facts, but they could hardly infer missing historical facts which are crucial for completing the overall knowledge landscape and providing more clues for predicting accurate future events.
These limit the practicability of TKG applications as almost all of the existing TKG reasoning methods are designed specifically to address either one setting.
Experiments with a novel pipeline setting intuitively reveal that even with the SOTA methods from both settings, the composed methods show a frustrating decrease in reasoning performance. More in-depth analysis can be found in Section \ref{sec:pipeline} and Appendix \ref{sec:detailed_pipeline}.
Therefore, the motivation of this work is to propose a unified method that can accommodate two types of reasoning settings, enabling temporal knowledge graph reasoning to be conducted simultaneously for both the interpolation and the extrapolation.

To this end, we take inspiration from recent neural \cite{XuCNL21,LiJLGGSWC21} and symbolic \cite{SunZMH021,LiuMHJT22} TKG reasoning approaches. Neural network approaches can perform effective reasoning but lack interpretation as they cannot provide explicit rules to explain the reasoning results, while symbolic reasoning approaches use logical symbols and rules to perform reasoning tasks but are not suitable for handling ambiguous and noisy data due to their strict matching and discrete logic operations used during rule searching \cite{ZhangCZKD21}.
In this paper, we propose a \textbf{T}emporal \textbf{PA}th based \textbf{R}easoning (TPAR) model with a neural-symbolic fashion applicable to both the interpolation and the extrapolation TKG Reasoning.
Specifically, we utilize the Bellman-Ford Shortest Path Algorithm \cite{Ford1956NETWORKFT,Bellman1958ONAR,2010Baras} and introduce a recursive encoding method to score the destination entities of various temporal paths, and then our TPAR performs symbolic reasoning with the help of the obtained scores.
It is noticeable that the neural-driven symbolic reasoning fashion we adopted is more robust to the uncertainty data compared to traditional pure symbolic reasoning methods, and comprehensible temporal paths with fine interpretability as well.
We summarize our main contributions as follows:
\begin{enumerate}
\item We propose an original unified Temporal PAth-based Reasoning (TPAR) model for both the interpolation and extrapolation reasoning settings. To the best of our knowledge, this is the first work to achieve the best of both worlds.

\item We develop a novel neural-driven symbolic reasoning fashion on various temporal paths to enhance both the robustness and interpretability of temporal knowledge reasoning.

\item Comprehensive experiments show that TPAR outperforms SOTA methods on the link prediction task for both the interpolation and the extrapolation settings with decent interpretability and robustness. An intriguing pipeline experiment is meticulously designed to demonstrate the strengths of TPAR in addressing the unified prediction through both settings.
\end{enumerate}

\section{Related Work}
\subsection{Static KG Reasoning}
Static KG reasoning methods can be summarized into three classes: the translation models \cite{bordes2013translating,DBLP:conf/aaai/WangZFC14,linlearning},
the semantic matching models \cite{Yang2015EmbeddingEA,Trouillon2016ComplexEF}, and the embedding-based models \cite{Dettmers2018Convolutional, Shang2019End}. Recently, R-GCN \cite{Schlichtkrull2018Modeling} and CompGCN \cite{VashishthSNT20} extended GCN to relation-aware GCN for KGs.

\subsection{TKG Reasoning}
\paragraph{Interpolation TKG Reasoning}
Many interpolation TKG reasoning methods \cite{leblay2018deriving,dasgupta2018hyte,garcia2018learning,xu2019temporal,LacroixOU20,JainRMC20,XuCNL21} are extended from static KGs to TKGs.
Besides, TeRo \cite{xu2020tero} represents each relation as dual complex embeddings and can handle the time intervals between relations.
RotateQVS \cite{ChenWLL22} represents temporal evolutions as rotations in quaternion vector space and can model various complex relational patterns. T-GAP \cite{JungJK21} performs path-based inference by propagating attention through the graph.

\paragraph{Extrapolation TKG Reasoning}
To predict future facts, many extrapolation TKG reasoning methods \cite{jin2020recurrent,LiJLGGSWC21,HanDMGT21,ZhuCFCZ21,SunZMH021,LiGJPL000GC22,LiS022} have been studied.
For better interpretability, TLogic \cite{LiuMHJT22} adopts temporary logical rules extracted via temporary random walks, while xERTE \cite{HanCMT21} utilizes an explainable subgraph and attention mechanism for predictions.
Besides, RPC \cite{0006MLLTWZL23} employs relational graph convolutional networks and gated recurrent units
to mine relational correlations and periodic patterns from temporal facts.

\subsection{Neural-Symbolic Reasoning}

Neural-symbolic reasoning aims to combine the strengths of both approaches, namely the ability of symbolic reasoning to perform logical inference and the ability of neural networks to perform learning from data \cite{ZhangCZKD21}.
Generally, it can be divided into three categories: (1) Symbolic-driven neural fashion \cite{GuoWWWG16,GuoWWWG18,abs-1903-03772}, which targets neural reasoning but leverages the logic rules to improve the embeddings. (2) Symbolic-driven probabilistic fashion \cite{RichardsonD06,Qu019,RaedtKT07}, which replaces the neural reasoning with a probabilistic framework, i.e., builds a probabilistic model to infer the answers. (3) Neural-driven symbolic fashion \cite{LaoC10,LaoSPC12,NeelakantanRM15}, which aims to infer rules by symbolic reasoning, and incorporates neural networks to deal with the uncertainty and ambiguity of data.
For this work, we adopt the neural-driven symbolic fashion for our TKG reasoning.

\section{Preliminaries}
\subsection{Problem Definition of TKG Reasoning}\label{sec:pre_tkgr}
A Temporal knowledge graph is a collection of millions of temporal facts, expressed as $\mathcal{G} \subset \mathcal{E} \times \mathcal{R} \times \mathcal{E} \times \mathcal{T}$. Here we use $\mathcal{E}$ to denote the set of entities, $\mathcal{R}$ to denote the set of relations, and $\mathcal{T}$ to denote the set of time stamps. Each temporal fact in $\mathcal{G}$ is formed as a quadruple $(e_s, r, e_o, t)$, where a relation $r \in \mathcal{R}$ holds between a subject entity $e_s\in \mathcal{E}$ and an object entity $e_o\in \mathcal{E}$ at the time $t$.
And for each quadruple $(e_s,r,e_o,t))$, an inverse quadruple $(e_o,r^{-1},e_s,t))$ is also added to the dataset.

Let $\mathcal{O} = \{(e_s, r, e_o, t) | t \in [t_0, T]\}$ represent the set of known facts that we can observe, and let $[t_0, T]$ denote the time interval we can access.
For a query $\bm q = (e_q, r_q, ?, t_q)$, TKG Reasoning aims to infer the missing object entity given the other three elements. As we have two distinctive TKG Reasoning settings: the interpolation (inferring a query where $t_0 \leq t_q \leq T$) and the extrapolation (predicting facts for $t_q > T$),
we use a unified concept of "temporal background" to represent all known temporal facts of existing TKG.

\subsection{Bellman-Fold Based Recursive Encoding}\label{sec:bf_based_encoding}
The Bellman-Ford Algorithm \cite{Ford1956NETWORKFT,Bellman1958ONAR,2010Baras} provides a recursive approach to search for the shortest path in a graph.\footnote{See Appendix \ref{sec:Bellman-Ford} for more details.}
Since TKGs are far from complete, we do not directly copy it for TKG Reasoning, but utilize a recursive encoding method for representations.

\begin{figure}[ht]
\centering
\includegraphics[width=0.6\columnwidth]{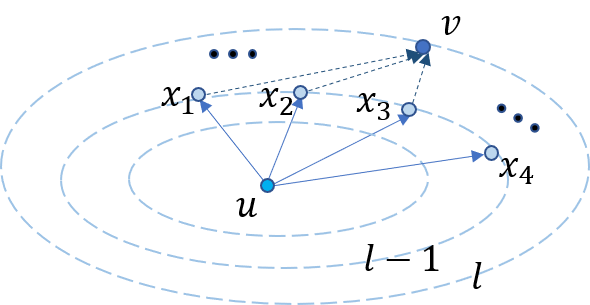}
\caption{An illustration of Bellman-Ford-based recursive encoding.}
\label{figure:recursive_encoding}
\end{figure}

 As illustrated in Fig.\ref{figure:recursive_encoding}, the basic idea of our Bellman-Ford-based recursive encoding is: once the representations of the $\ell-1$ iterations (the node $x$) starting from node $u$ are ready, we can obtain the representations of the $\ell$ iterations  (the node $v$) by combining them with the edge $(x,v)$:

\begin{footnotesize}
\begin{equation}\label{eq:boundary}
    \bm h^{0}(u, v) \leftarrow \mathbbm{1}(u = v) 
    \nonumber
\end{equation}  
\end{footnotesize}
\vspace{-0.1in}
\begin{scriptsize}
\begin{equation}
\label{eq:bf_iteration}
    \bm h^{\ell}(u, v) \leftarrow   
     \left(\bigoplus_{(x, v)\in\mathcal{E}(v)} \bm h^{\ell-1}(u, x) \otimes w(x, v)\right) \oplus \bm h^{(0)}(u, v)\,,
     \nonumber
\end{equation}
\end{scriptsize}
where $\mathbbm{1}(u = v)$ is the indicator function that outputs 1 if $u=v$ and 0 otherwise, $\otimes$ is the multiplication operator, $\oplus$ is the summation operator,
$\mathcal{E}(v)$ is the set of all edges connected to the node $v$, and $w(x,v)$ is the weight of the edge $(x,v)$. 

\section{Methodology}
In this section, we introduce a novel temporal path-based reasoning model with a neural-driven symbolic fashion for both the interpolation and the extrapolation TKG Reasoning settings.

\subsection{Temporal Path in TKGs}\label{section:path}
\begin{myDef}
	\label{definition:link}
   (Temporal link). In a TKG, each entity is viewed as a node in the graph, and a temporal link is viewed as an edge connecting two nodes which represent the subject $e_s$ and the object entity $e_o$ in a certain quadruple $(e_s, r, e_o, t)$. Due to the heterogeneity and time constraints, each link is bonded with a relation $r$ and a particular time $t$. Thus, we can denote a temporal link as $\small{\begin{matrix} t \\ r \end{matrix} (e_s, e_o)}$. 
   
\end{myDef}

\begin{myDef}
	\label{definition:path}
   (Temporal path). A temporal path $\mathcal{P}$ is a combination of several temporal links that are sequentially connected from subject entity to object entity. We denote a temporal path $\mathcal{P}$ as
   \begin{equation}\label{equation:path}
   \small{
    \begin{matrix} t_1 \\ r_1 \end{matrix} (e_1, e_2) \wedge \begin{matrix} t_2 \\ r_2 \end{matrix} (e_2, e_3) \wedge \cdot \cdot \cdot \begin{matrix} t_{\ell} \\ r_{\ell} \end{matrix} (e_{\ell}, e_{\ell+1})\,, }
\end{equation}
or $ \small{\bigwedge_{i=1}^{\ell}  \begin{matrix} t_i \\ r_i \end{matrix} (e_i, e_{i+1}) }$ for short, where $e_1$, $e_2$, $\cdot \cdot \cdot e_{l+1}$ are sequentially connected to form a chain, and $\ell$ is the path length.
We take $e_1$ and $e_{l+1}$ as the origin and the destination of the path respectively, and use $\mathcal{P}_{(i)}$ to denote the temporal link $\small{ \begin{matrix} t_i \\ r_i \end{matrix} (e_i, e_{i+1}) }$ of the $i$-th step in $\mathcal{P}$.\footnote{An illustration of temporal paths and temporal links is shared in Appendix \ref{sec:illustraion}.}
\end{myDef}

For the extrapolation setting, a non-increasing temporal path fashion has been applied in recent works \cite{HanCMT21,LiuMHJT22}, and requires $t_1 \geq t_2 \geq \cdot \cdot \cdot t_{\ell} \geq t_{\ell+1}$. In this paper, we are not going to follow the non-increasing setup. For a query $\bm q = (e_q, r_q, ?, t_q)$, we require each $t_i$ in a temporal path $\mathcal{P}$ to satisfy $t_i < t_q$ such that we only use information from the past to predict the future, but have no restrictions to the chronological orders among $t_1$, $t_2$, $\cdot \cdot \cdot$ $t_l$, i.e. we can have $t_i < t_{i+1}$, $t_i = t_{i+1}$ or $t_i > t_{i+1}$.
This relaxation provides more temporal paths deriving from the temporal background, enabling us to utilize more information for reasoning. Further discussions for this can be found in Appendix \ref{sec:ablation_time_constrain}.

For the interpolation setting, we mainly focus on how to find the missing object entity from the given subject entity $e_q$ and indeed we ignore any time constraints. For a temporal link $\small{\begin{matrix} t_i \\ r_i \end{matrix} (e_i, e_{i+1})}$ in the temporal background, whether it is before or after $t_q$, we can include it in a temporal path as long as it has been known to us.

\subsection{Relative Time Encoding on Temporal Links}\label{section:relative_time}
We adopt relative time encoding and focus on the impact of a certain temporal link in the temporal path on a query.
For each temporal link $\begin{matrix} t_i \\ r_i \end{matrix} (e_i, e_{i+1})$, we first obtain the relative time $\Delta t_i = t_i - t_q$. Notice that a temporal link itself is query-independent, but the obtained relative time $\Delta t_i$ is query-dependent. For the extrapolation setting, there is always an inequality $\Delta t_i < 0$ since $t_i < t_q$. But for the interpolation setting, $\Delta t_i < 0$, $\Delta t_i = 0$ and $\Delta t_i > 0$ are all possible since there is no restriction on the order of $t_i$ and $t_q$.

As mentioned by \citet{ZhuCFCZ21}, many facts have occurred repeatedly throughout history, such as economic crises that occur every seven to ten years \cite{AndreyEconomicCrisis}. We pay close attention to the periodic property as well as the non-periodic property of time. We then extend the idea of periodic and non-periodic time vectors \cite{LiS022} to a relative time encoding fashion:
\begin{equation}\label{equation:time_np}
   \small{
\begin{aligned}
 \bm{h}^p_{t_i} = \sin(\omega_p \Delta t_i + \varphi_p )
\\
\bm{h}^{np}_{t_i} = \omega_{np} \Delta t_i + \varphi_{np} \,,
\end{aligned}}
\end{equation}
where $\omega_p$, $\varphi_p$, $\omega_{np}$, $\varphi_{np}$ are learnable parameters, $\sin()$ denotes the sine function, and the obtained $\bm{h}^p_{t_i}$ and $\bm{h}^{np}_{t_i}$ are the periodic and the non-periodic relative time vectors, respectively.
For the periodic vector $\bm{h}^p_{t_i}$, the frequency $\omega_p$ and the phase shift $\varphi_p$ reflect the periodic property. For the non-periodic vector $\bm{h}^{np}_{t_i}$, $\omega_{np}$ acts as a velocity vector \cite{HanCMT20} indicating the temporal evolving dynamics.

Finally, we combine both periodic and non-periodic properties to design our relative time embedding as:
\begin{equation}\label{equation:time_encoding}
\small{
    \bm{h}_{t_i} = \bm{h}^p_{t_i} + \bm{h}^{np}_{t_i} \,,}
\end{equation}
where we use the same embedding dimension $d$ for $\bm{h}^p_{t_i}$, $\bm{h}^{np}_{t_i}$ and $\bm{h}_{t_i}$.

\subsection{Recursive Encoding on Temporal Paths}\label{sec:recursive_encoding}
In Section \ref{sec:bf_based_encoding}, we have mentioned the idea of Bellman-Ford-based recursive encoding. Now, we start from the query and introduce our recursive encoding on temporal paths for TKG reasoning.

\paragraph{Basic Design}
Our recursive encoding on temporal paths aims to encode the destination entities of various temporal paths. And the basic idea is: Once the representations of the destination entities of all the $(\ell-1)$-length paths are ready, we can combine them with the link $\mathcal{P}_{(l)}$ to encode the destination entities of the $\ell$-length paths.

For a query $\bm q = (e_q, r_q, ?, t_q)$, our recursive encoding is query-specific, and we start with the query entity $e_q$.
Initially, we can only reach the original entity $e_q$, so we initialize $\bm{h}_{e_q}^{0}=\bm0$ and the entity set $\mathcal E_{e_q}^{0}=\{e_q\}$. In the $\ell$-th step, we collect all the $\ell$-length temporal paths from temporal background based on $e_q$, and update $\mathcal E_{e_q}^{\ell}$ with the destination entities. We utilize $\mathcal{D}(\mathcal{P}_{e_q})$ to denote the destination entity set of the temporal paths $\mathcal{P}_{e_q}$, and $\mathcal{P}_{e_q,(\ell)}$ to denote the set of links of the $\ell$-th step in $\mathcal{P}_{e_q}$. Thus, we actually have $\mathcal E_{e_q}^{\ell} = \mathcal E_{e_q}^{\ell -1} \cup \mathcal{D}(\mathcal{P}_{e_q})$ for $\ell > 1$.

Then, the message is passed along temporal paths $\mathcal{P}_{e_q}$ from $ e_s$ to $ e_o$, where $\small{\begin{matrix} t \\ r \end{matrix} (e_s, e_o)} \in  \mathcal{P}_{e_q,(l)} $. And we can get the representations for all $e_o \in \mathcal{D}(\mathcal{P}_{e_q})$ for $\ell = 1, 2, \cdots L $ by:
\begin{equation}\label{eq:encooding}
\small{
  \bm{h}_{e_o}^{\ell}=
 \delta
		    \Big(\bm{W}^{\ell} \sum\nolimits_{  \mathcal{P}_{e_q,(\ell)}} \phi\big(\bm h^{\ell\!-\!1}_{e_s}, \bm h_r^\ell, \bm{h}_{t} \big)\Big) \,,}
\end{equation}
where $\delta$ denotes an activation function, $\bm{W}^{\ell} \in \mathbb{R}^{d \times d}$ is a learnable parameter, $\phi ()$ denotes a function to compute and process the messages, $\bm h_r^\ell$ is the embedding for relation $r$ in the $\ell$-th step, and $\bm{h}_{t}$ is a result of the relative time encoding that we have introduced in Section \ref{section:relative_time}.
The proposed recursive encoding algorithm is shown in Algorithm \ref{alg:Encoding}.\footnote{The time complexity is $\mathcal{O}(|\mathcal{E}|d + |\mathcal{V}| d^2)$, where $|\mathcal{E}|$ and $|\mathcal{V}|$ denote the number of entities and relations respectively.}

\begin{center}  
\begin{minipage}{0.96\columnwidth}  
\begin{algorithm}[H]
\small
	\caption{\small{The Proposed Recursive Encoding Algorithm}}
	\label{alg:Encoding}
	\renewcommand{\algorithmicrequire}{\textbf{Input:}}
	\renewcommand{\algorithmicensure}{\textbf{Output:}}
	\begin{algorithmic}[1]
		\REQUIRE A temporal knowledge graph $G$, a query $\bm q = (e_q, r_q, ?, t_q)$, the maximum length $L$ 
		\ENSURE The query-specific entity representations    
        \STATE initialize $\bm{h}_{e_q}^{0}\!=\bm0$ and the entity set $\mathcal E_{e_q}^{0}=\{e_q\}$;
        \FOR{$\ell=1,2,\cdots L$}
		    \STATE collect all the $\ell$-length temporal paths and update $\mathcal E_{e_q}^{\ell}$ with the destination entity set $\mathcal{D}(\mathcal{P}_{e_q})$;

            \STATE message passing along temporal paths $\mathcal{P}_{e_q}$ from $ e_s \in \mathcal{D}(\mathcal{P}_{e_q,{(\ell-1)}})$ to $ e_o \in \mathcal{D}(\mathcal{P}_{e_q,{(\ell)}})$: \\
            \small{
        $\bm{h}_{e_o}^{\ell} = \delta
		    \Big(\bm{W}^{\ell} \sum\nolimits_{ \mathcal{P}_{e_q,(l)} } \phi\big(\bm h^{\ell\!-\!1}_{e_s}, \bm h_r^\ell, \bm{h}_{t} \big)\Big)$  },\\
        \ENDFOR
        \STATE assign $\bm h_{e_a}^L=\bm 0$ for all $e_a\notin \mathcal E_{e_q}^L$;
		\RETURN $\bm h_{e_a}^L$ for all $e_a\in\mathcal{E}$.
	\end{algorithmic}
\end{algorithm}
\end{minipage}
\end{center}

\paragraph{Message Passing}
We then give a detailed description of the message passing along temporal paths from $ e_s$ to $ e_o $.

Given a quadruple $(e_s,r,e_o,t)$ in temporal ground, we can transform it to a temporal link $\small{\begin{matrix} t \\ r \end{matrix} (e_s, e_o)} $.
We take the message as a combination of the query-specific representation of $e_s$ learned in the $(\ell-1)$-th recursive encoding step, the representation of $r$, and the representation of $t$:
\begin{equation}
\small{
\bm {M}_{(e_s, r, e_o, t)|\bm q} = \bm h^{\ell\!-\!1}_{e_s} + \bm h_r^\ell + \bm{h}_{t} \,,}
\end{equation}
Note that we use the same embedding dimension $d$ for $\bm h^{\ell\!-\!1}_{e_s} $, $\bm h_r^\ell$, $\bm h_{r_q}^\ell$ and $\bm{h}_{t} $.

Then, we utilize an attention-based mechanism to weigh the message. 
Inspired by graph attention networks \cite{VelickovicCCRLB18}, we first take a concatenation operation on $\bm h^{\ell\!-\!1}_{e_s}$, $\bm h_r^\ell$, $\bm h_{r_q}^\ell$ and $\bm{h}_{t} $ to get
\begin{equation}\label{equation:unit}
\small{
\bm {U}_{(e_s, r, e_o, t)|\bm q} = concat[\bm h^{\ell\!-\!1}_{e_s}(e_q,r_q,t_q) || \bm h_r^\ell || \bm h_{r_q}^\ell || \bm{h}_{t}]
\,, }
\end{equation}
where $\bm {U}_{(e_s, r, e_o, t)|\bm q} \in \mathbb{R}^{4d}$ is the concatenated result.
Then, the attention weight for the message can be defined as
\begin{equation}\label{equation:attention}
\small{
\alpha^{\ell}_{(e_s, r, e_o, t)|\bm q} = \sigma \Big((\bm{w}^{\ell}_{\alpha})^T ReLU (\bm{W}^{\ell}_{\alpha} \cdot \bm {U}_{(e_s, r, e_o, t)|\bm q} ) \Big) 
\,,}
\end{equation}
where $\bm{w}^{\ell}_{\alpha} \in \mathbb{R}^{d_a}$, $\bm{W}^{\ell}_{\alpha} \in \mathbb{R}^{d_a \times 4d}$ are learnable parameters for attention ($d_a$ is the dimension for attention).
Just like \cite{ZhangY22}, we use the Sigmoid function $\sigma$ rather than the Softmax function to ensure multiple temporal links can be selected in the same recursive step.

\paragraph{Aggregation}
With messages and the weights obtained, we take an aggregation operation on messages along different temporal links in various temporal paths, by means of a weighted sum of messages. For the $\ell$-th recursive step, our encoded representation is specified as
\begin{equation}
\small{
\bm{h}_{e_o}^{\ell} = \delta
		    \Big(\bm{W}^{\ell} \sum\nolimits_{  \mathcal{P}_{e_q,(\ell)} } \alpha^{\ell}_{(e_s, r, e_o, t)|\bm q} \cdot \bm {M}_{(e_s, r, e_o, t)|\bm q}
      \Big) \,. }
\end{equation}

So far, we have provided a detailed version of our recursive encoding function (also see a general version in Equation \ref{eq:encooding}) for each recursive step $\ell =1,2,\cdots L $.
After $L$ recursive steps, the representation of each entity inside various $L$-length temporal paths (starting from $e_q$) can be learned step by step.
We assign $\bm h_{e_a}^L\!(e_q,r_q,t_q)=\bm 0$ for all $e_a\notin \mathcal E_{e_q}^L$.
For each entity $e_a \in \mathcal E$, we finally get its representation $\bm{h}_{e_a}^{L}$ after $L$-length recursive encoding process.

\subsection{Reasoning on Temporal Paths}
Then, we can leverage the learned representations to measure the quality of temporal paths to reason.

\paragraph{Neural based Scoring}
To accomplish neural-symbolic reasoning, we take a neural-based scoring first and use a simple scoring function to compute the likelihood for each candidate answer entity $e_a$:
\begin{equation}\label{eq:score_function}
\small{
f\left(\bm q, e_{a}\right)  = \bm{w}^T \bm{h}_{e_a}^{L}(e_q, r_q, t_q) \,,
}
\end{equation}
where $\bm{w} \in \mathbb{R}^d$ is a parameter for scoring.

And for training loss, we use a multi-class log-loss function to train the neural networks for our TKG Reasoning, which has been proven effective \cite{LacroixUO18,ZhangY22}:
\begin{equation}
\label{eq:loss_function}
\small{
\mathcal{L}  = \sum_{\small{\left(\bm q, e_{a}\right) \in \mathcal{T}_{\text {train}}}}\left(-f\left(\bm q, e_{a}\right)+\log \left(\sum_{\forall e \in \mathcal{E}} e^{f\left(\bm q, e\right)}\right)\right), 
}
\end{equation}
where $\mathcal{T}_{\text {train}}$ is set of training queries, and $\sum_{\forall e \in \mathcal{E}} $ is for all the quadruples w.r.t. the same query $\bm q$.

\paragraph{Symbolic Reasoning}
As for our symbolic reasoning, we target both the suitable temporal path and its destination entity.
Using $m_i$ to denote the $i$-th intermediary entity, our reasoning can be expressed as
\begin{equation}
\small{
\begin{aligned}
\small{\begin{matrix} t_1 \\ r_1 \end{matrix} (e_q, m_1)} & \wedge \small{\begin{matrix} t_2 \\ r_2 \end{matrix} (m_1, m_2)} \wedge \cdot \cdot \cdot \small{\begin{matrix} t_{L} \\ r_{L} \end{matrix} (m_{L-1}, e_{a})}
    \\
  & \to \small{\begin{matrix} t_q \\ r_q \end{matrix} (e_q, e_a)}
    \,,
\end{aligned}
}
\end{equation}
where the left part of $\to$ denotes a temporal path composed of $L$ sequentially connected temporal links, and the right part is the link we aim to infer.

We use $\bm A_{e_q}^{\ell} \in {\{0,1\}}^{N_{\ell} \times N_{\ell}}$ to denote the adjacent matrix of the $\ell$-th step in the set of $L$-length paths $\mathcal{P}_{e_q}$ starting from $e_q$, i.e. $ \mathcal{P}_{e_q,(\ell)} $, where $N_{\ell}$ is the number of entities involved for $\ell =1,2, \cdots L$.
If $\small{ \begin{matrix} t_1 \\ r_1 \end{matrix} (e_q, m_1)} \wedge \small{\begin{matrix} t_2 \\ r_2 \end{matrix} (m_1, m_2)} \wedge \cdot \cdot \cdot \small{\begin{matrix} t_{L} \\ r_{L} \end{matrix} (m_{L-1}, e_{a})} \to \small{\begin{matrix} t_q \\ r_q \end{matrix} (e_q, e_a)}$ holds, ideally there exists a suitable path $\mathcal{P}^{\star}$:
\begin{equation}
\small{
 \mathcal{P}^{\star} =  \beta^1  \bm A_{e_q}^1 	\otimes \beta^2 \bm A_{e_q}^2 \cdots  \otimes \beta^L \bm A_{e_q}^L \,. }
\end{equation}
where $\small{\beta^{\ell} \in {\{0,1 \}}^{N_{\ell} \times N_{\ell}}}$ is the choice of links for step $\ell$.

From Equation \ref{equation:unit} and \ref{equation:attention}, we can know the learned $\small{\alpha^{\ell}_{(e_s, r, e_o, t)|\bm q}  = {MLP}^{\ell}(e_s,r,e_q,r_q,\Delta t)}$, and abbreviate it as $\small{\alpha^{\ell}} $.
Our recursive encoding provides
\begin{equation}
\small{
\underset{\bm \Theta}{\arg\max} \,\, Score(\mathcal{P}) =\alpha^{1}  \bm A_{e_q}^1 	\otimes \alpha^{2} \bm A_{e_q}^2 \cdots  \otimes \alpha^{L} \bm A_{e_q}^L \,,
}
\end{equation}
where $\small{Score(\mathcal{P})}$ is the score of the path $P$ obtained by Equation \ref{eq:score_function}.
Based on universal approximation theorem \cite{HORNIK1991251}, there exists a set of parameters $\Theta$ that can learn $\alpha^{\ell} \simeq \beta^{\ell}$, to make $\mathcal{P} \simeq \mathcal{P}^{\star}$.
Note that multiple temporal paths may lead to the same destination. In such cases, the score can be regarded as a comprehensive measure of the various reasoning paths. Case-based analysis on this can be found in Section \ref{sec:case_study}.

To summarize, our temporal path-based reasoning first collects temporal paths from the temporal background, then scores the destination entities of various temporal paths by attention-based recursive encoding, and lastly takes the entity with the highest score as the reasoning answer.

\section{Experiments}

\subsection{Evaluation Protocol}
Link prediction task that aims to infer incomplete time-wise fact with a missing entity ($(s, r, ?, t)$ or $(?, r, o, t)$) is adopted to evaluate the proposed model for both interpolation and extrapolation reasoning over TKGs.
During inference, we follow the same procedure of \citet{xu2020tero} to generate candidates. For a test sample $(s, r, o, t)$, we first generate candidate quadruples set $C = \{(s, r, \overline{o}, t): \overline{o} \in \mathcal{E}\} \cup \{(\overline{s}, r, o, t): \overline{s} \in \mathcal{E}\}$ by replacing $s$ or $o$ with all possible entities, and then rank all the quadruples by their scores (Equation~\ref{eq:score_function}).
We use the time-wise filtered setting \cite{xu2019temporal,goel2020diachronic} to report the experimental results.

We use {ICEWS14}, {ICEWS05-15} \cite{garcia2018learning}, {YAGO11k} \cite{dasgupta2018hyte} and {WIKIDATA12k} \cite{dasgupta2018hyte} datasets for interpolation reasoning evaluation, and use {ICEWS14} \cite{garcia2018learning}, {ICEWS18} \cite{jin2020recurrent}, {YAGO} \cite{Mahdisoltani2015YAGO3AK} and {WIKI} \cite{leblay2018deriving} for extrapolation.\footnote{More details for datasets can be found in Appdendix \ref{sec:app_datasets}.}
The performance is reported on the standard evaluation metrics: the proportion of correct triples ranked in the top 1, 3 and 10 (Hits@1, Hits@3, and Hits@10), and Mean Reciprocal Rank (MRR). All of them are the higher the better.
For all experiments, we report averaged results across 5 runs, and we omit the variance as it is generally low.

\subsection{Main Results}
\subsubsection{Interpolation Reasoning}\label{sec:exper_inter}
\begin{table*}[!ht]
 \renewcommand\arraystretch{1.1}
\centering
\begin{minipage}[t]{0.99\textwidth}
\centering
\resizebox{0.99\textwidth}{!}{

\begin{tabular}{l|cccc|cccc|cccc|cccc}
\toprule
\multirow{2}{*}{Interpolation}
  & \multicolumn{4}{c|}{ICEWS14}   & \multicolumn{4}{c|}{YAGO11k}  & \multicolumn{4}{c|}{ICEWS05-15} & \multicolumn{4}{c}{WIKIDATA12k} \\
\cline{2-17}
    & Hits@1        & Hits@3  & Hits@10  & MRR  & Hits@1           & Hits@3  & Hits@10 & MRR  & Hits@1        & Hits@3  & Hits@10  & MRR & Hits@1        & Hits@3  & Hits@10  & MRR\\ \hline
 TTransE   & 7.4        & - & 60.1 & 25.5  & 2.0 & 15.0 & 25.1 & 10.8 &8.4 &- &61.6 &27.1  &9.6 &18.4 &32.9 &17.2\\
HyTE      & 10.8        & 41.6 & 65.5 & 29.7 & 1.5  & 14.3 & 27.2 & 10.5  &11.6 &44.5 &68.1 &31.6 &9.8 &19.7 &33.3 &18.0 \\
TA-DistMult & 36.3        & - & 68.6 & 47.7   &10.3 &17.1 &29.2 & 16.1 &34.6 &- &72.8 &47.4 &12.2 &23.2 &44.7 &21.8\\
ATiSE    &43.6       & 62.9 & 75.0 & 55.0 & 11.0  &17.1 & 28.8 & 17.0 &37.8 &60.6 &79.4 &51.9 &17.5 &31.7 &48.1 &28.0
 \\
TeRo     &  46.8       & 62.1 & 73.2 & 56.2 & 12.1 &19.7 &31.9 &18.7
&46.9 &66.8 &79.5 &58.6 &19.8 &32.9 &50.7 &29.9 \\ 
TComplEx  &53 & 66 &77 &61 & 12.7 & 18.3 & 30.7 & 18.5 
&59 &71 &80 &66 &23.3 &35.7 &53.9 &33.1
\\
T-GAP  &  50.9       & 67.7 & 79.0 & 61.0 & - &- &- &-
&  56.8       & 74.3 & 84.5 & 67.0 & - &- &- &-
\\ 
TELM  &  54.5       & 67.3 & 77.4 & 62.5 & 12.9 &19.4 &32.1 &19.1
&  59.9 &72.8 &82.3 &67.8 &23.1 &36.0 &54.2 &33.2\\ 
RotateQVS   & 50.7        & 64.2 &75.4 &59.1  &12.4 & 19.9 & 32.3 & 18.9 & 52.9 &70.9 &81.3 &63.3  &20.1 &32.7 &51.5 &30.2
 \\
TGoemE++ &54.6 &68.0&78.0&62.9 &13.0&19.6&32.6&19.5 &60.5&73.6&83.3&68.6 &23.2&36.2&54.6&33.3
 \\
\hline
TPAR (Ours) & \bfseries 57.03       &  \bfseries 69.74 & \bfseries 80.41 & \bfseries 65.07 & \bfseries 17.35 & \bfseries 25.14 &\bfseries 37.42 &\bfseries 24.12  & \bfseries 62.17       &\bfseries 75.29   &\bfseries 85.86  &\bfseries 69.33  & \bfseries 25.05 & \bfseries 38.68 &\bfseries 54.79 &\bfseries 34.89
\\ 

\bottomrule
\end{tabular}

}
\end{minipage}
\caption{\textbf{Interpolation} TKG Reasoning results (in percentage) on link prediction over four experimented datasets.}
\label{table:results_neicha}
\end{table*}
\begin{table*}[!ht]
 \renewcommand\arraystretch{1.1}
\centering
\begin{minipage}[t]{0.99\textwidth}
\centering
\resizebox{0.99\textwidth}{!}{

\begin{tabular}{l|cccc|cccc|cccc|cccc}
\toprule
\multirow{2}{*}{Extrapolation}
  & \multicolumn{4}{c|}{ICEWS14}   & \multicolumn{4}{c|}{YAGO}  & \multicolumn{4}{c|}{ICEWS18} & \multicolumn{4}{c}{WIKI} \\
\cline{2-17}
    & Hits@1        & Hits@3  & Hits@10  & MRR  & Hits@1           & Hits@3  & Hits@10 & MRR  & Hits@1        & Hits@3  & Hits@10  & MRR  & Hits@1           & Hits@3  & Hits@10 & MRR \\ \hline
 RE-NET &30.11 &44.02 &58.21 &39.86    &58.59 &71.48 &86.84  &66.93  &19.73 &32.55 &48.46 &29.78 &50.01 &61.23 &73.57 &58.32
\\
CyGNeT   &27.43 &42.63 &57.90 &37.65  &58.97 &76.80 &86.98 &68.98
&17.21 &30.97 &46.85 &27.12 &47.89 &66.44 &78.70 &58.78\\
TANGO &- &- &- &-  &60.04 &65.19 &68.79 &63.34
 &18.68 &30.86 &44.94 &27.56 &51.52 &53.84 &55.46 &53.04 \\
xERTE  &32.70 &45.67 &57.30 &40.79 &80.09 &88.02 &89.78 &84.19
&21.03 &33.51 &46.48 &29.31 &69.05 &78.03 &79.73 &73.60\\
RE-GCN  &31.63 &47.20 &61.65 &42.00 &78.83 &84.27 &88.58 &82.30
&22.39 &36.79 &52.68 &32.62  &74.50 &81.59 &84.70 &78.53\\
TITer  &32.76 &46.46 &58.44 &41.73 &80.09 &89.96 &90.27 &87.47
&22.05 &33.46 &44.83 &29.98  &71.70 &75.41 &76.96 &73.91 \\
CEN   &32.08 &47.46 &61.31 &42.20  &- &- &- &-
&21.70 &35.44 &50.59 &31.50 &75.05 &81.90 &84.90 &78.93\\
TLogic  &33.56 &48.27 &61.23 &43.04 &- &- &- &-
&20.54 &33.95 &48.53 &29.82 &- &- &- &-\\
TIRGN    &33.83 &48.95 &63.84 &44.04  &84.34 &91.37 &92.92 &87.95 &23.19 &37.99 &54.22 &33.66 &77.77 &85.12 &87.08 &81.65\\
RPC   &34.87 &49.80 &65.08 &44.55  &85.10 &92.57 &94.04 &88.87 &24.34 &38.74 &55.89 &34.91  &76.28 &85.43 &88.71 &81.18\\
\hline
TPAR (Ours) & \bfseries 36.88       &  \bfseries 52.28 & \bfseries 65.89 & \bfseries 46.89 
&\bfseries 89.67    &\bfseries 92.93 &\bfseries 94.60 &\bfseries 91.53 &\bfseries 26.58 &\bfseries 39.27 &\bfseries 56.94 &\bfseries 35.76 &\bfseries 79.85 &\bfseries 87.04 &\bfseries 88.96 &\bfseries 83.20
\\ 
\bottomrule
\end{tabular}

}
\end{minipage}
\caption{\textbf{Extrapolation} TKG Reasoning results (in percentage) on link prediction over four experimented datasets.}
\label{table:results_waitui}
\end{table*}
Due to space limitations, we share the experimental details in Appendix \ref{sec:appendix_inter}.
Table \ref{table:results_neicha} shows the results of the interpolation TKG Reasoning on link prediction over four experimented datasets where the proposed TPAR continuously outperforms all baselines across all metrics.
We specifically compare our proposed TPAR with another GNN-based method, T-GAP \cite{JungJK21}, which samples a subgraph from the whole TKG for each node. In contrast, our TPAR recursively traverses the entire graph based on previously visited nodes to collect temporal paths. Our experimental results show that TPAR outperforms T-GAP across all evaluation metrics on the ICEWS14 and ICEWS05-15 datasets.
This can be attributed to TPAR's ability to capture the temporal information of the entire graph and generate longer paths, which leads to more accurate and comprehensive TKG reasoning.

\subsubsection{Extrapolation Reasoning}\label{sec:exper_extra}
Experimental details for Extrapolation reasoning are shown in Appendix \ref{sec:appendix_extra}.
Table \ref{table:results_waitui} presents the results of the extrapolation TKG reasoning on link prediction across four different datasets.
Notably, we compare TPAR with two SOTA methods, namely, the neural network-based RPC \cite{0006MLLTWZL23} and the symbolic-based TLogic \cite{LiuMHJT22}. Our results show that TPAR outperforms all baselines, underscoring the benefits of combining neural and symbolic-based approaches for TKG reasoning.

\subsection{Analysis on Pipeline Setting}\label{sec:pipeline}
To test the hypothesis that completing missing knowledge about the past can enhance the accuracy of predicting future knowledge, we design a novel experimental pipeline.
Building upon the facts within the known time range, we sample a ratio of facts to train interpolation, while the other unsampled facts have either the subject entity or the object entity masked, forming incomplete quadruples.
Specifically, we have three settings: set the ratio to 60\% for Setting A, 70\% for Setting B, and 80\% for Setting C.\footnote{See Appendix \ref{sec:detailed_pipeline} for full pipeline setting details.} The lower the ratio, the more incomplete facts need to be interpolated first.

The pipeline involves first completing these incomplete quadruples through interpolation, and subsequently predicting future events.
\textbf{Baselines}: Since there are currently no existing methods capable of both interpolation and extrapolation, we assemble SOTA interpolation and extrapolation methods\footnote{For fairness, we do not use TGoemE++ and RPC as they have not made their code publicly available.}, and get two combinations: a) TELM and TLogic, b) RotateQVS and TIRGN.

\begin{table}[t]
 \renewcommand\arraystretch{1.3}
\centering
\begin{minipage}[t]{0.99\columnwidth}
\centering
\resizebox{0.99\columnwidth}{!}{

\begin{tabular}{c|c|c|c|c}
\toprule
\multirow{2}{*}{Interpolation} & \multirow{2}{*}{Extrapolation}   & \multicolumn{3}{c}{Setting} 
\\ \cline{3-5}
 &  & \textbf{A} (Ratio: 60\%) & \textbf{B} (Ratio: 70\%) & \textbf{C} (Ratio: 80\%)
\\ \Xhline{1px}
  TELM & TLogic & 40.88 & 41.55 & 42.09
   \\ \hline
 \XSolidBold & TLogic & 41.85 & 41.88 & 42.18 
   \\  \Xhline{1px}
 RotateQVS & TIRGN & 40.86 & 42.35 & 42.78
   \\ \hline
 \XSolidBold & TIRGN & 42.53 & 43.68 & 43.69
   \\ \Xhline{1px}
 TPAR & TPAR & 44.56 & 46.07 & 46.66
   \\ \hline
\XSolidBold & TPAR & 43.86 & 45.49 & 36.36
   \\
\bottomrule
\end{tabular}

}
\end{minipage}
\caption{MRR performance of the pipeline setting on ICEWS 14.}
\label{table:pipeline}
\end{table}

The experimental results in Table \ref{table:pipeline}\footnote{See Table \ref{table:pipeline_full} for the more detailed results.} show two key observations:
a) For both of the SOTA combination baselines, employing the pipeline (interpolation followed by extrapolation) leads to a reduction in extrapolation performance compared to straightforward extrapolation.
And this reduction in the effect is gradually alleviated as the ratio increases from 60\% to 80\%. We believe that this could be attributed to the fact that the newly acquired knowledge from interpolation may contain errors, which have the potential to propagate and amplify in subsequent extrapolation.
As the amount of newly acquired knowledge decreases, the likelihood of introducing errors into the system also diminishes.
b) In all three settings, our TPAR with pipeline outperforms the straightforward extrapolation results, which demonstrates the superiority of our TPAR model to effectively integrate interpolation and extrapolation as a unified task for TKG reasoning.

\subsection{Case Study: Path Interpretations}\label{sec:case_study}

Benefiting from the neural-symbolic fashion, we can interpret the reasoning results through temporal paths. Following local interpretation methods \cite{ZeilerF14,ZhuZXT21}, we approximate the local landscape of our TPAR with a linear model over the set of all paths, i.e., 1st-order Taylor polynomial.
Thus, the importance of a path can be defined as its weight in the linear model and can be computed by the partial derivative of the prediction w.r.t. the path.
Formally, top-k temporal path interpretations for $f(\bm q,e_a)$ are defined as
\begin{equation}
\small{
    P_1,P_2,...,P_k =  
    \underset{P \in \mathcal{P}_{e_q e_a}}{top\text{-}k}
\frac{\partial{f(\bm q, e_a)}}{\partial{P}}\,,}
\end{equation}
where $\mathcal{P}_{e_q e_a}$ denotes the set of all temporal paths starting from $e_q$ to $e_a$.

\begin{figure}[t]
\centering
    \subfigure[\textbf{Interpolation} for the maximum length $L=2$.]{
       \centering
        \includegraphics[width=0.85\linewidth,height=2.5cm]{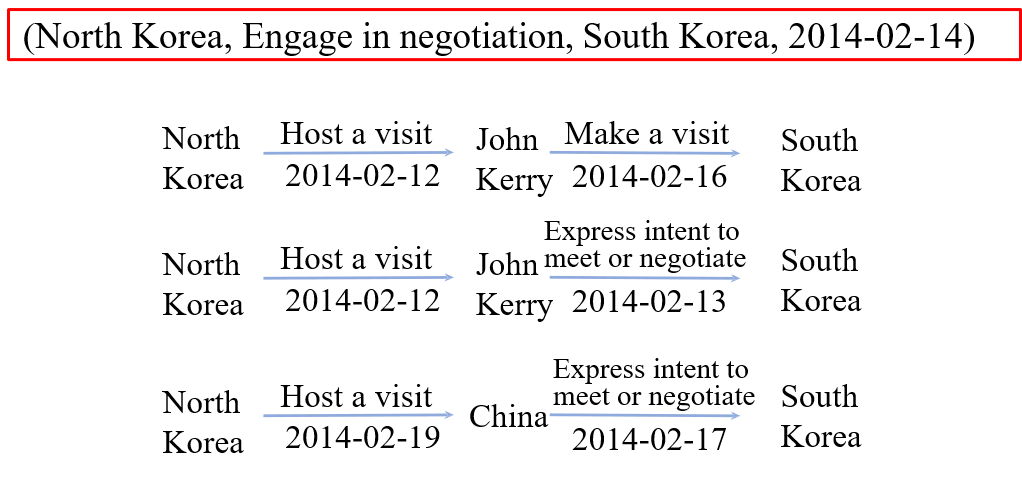}
        \label{fig:case_intro}
    }
    \\
    \subfigure[\textbf{Extrapolation} for the maximum length $L=3$.]{
        \centering
        \includegraphics[width=0.98\linewidth,height=6cm]
        {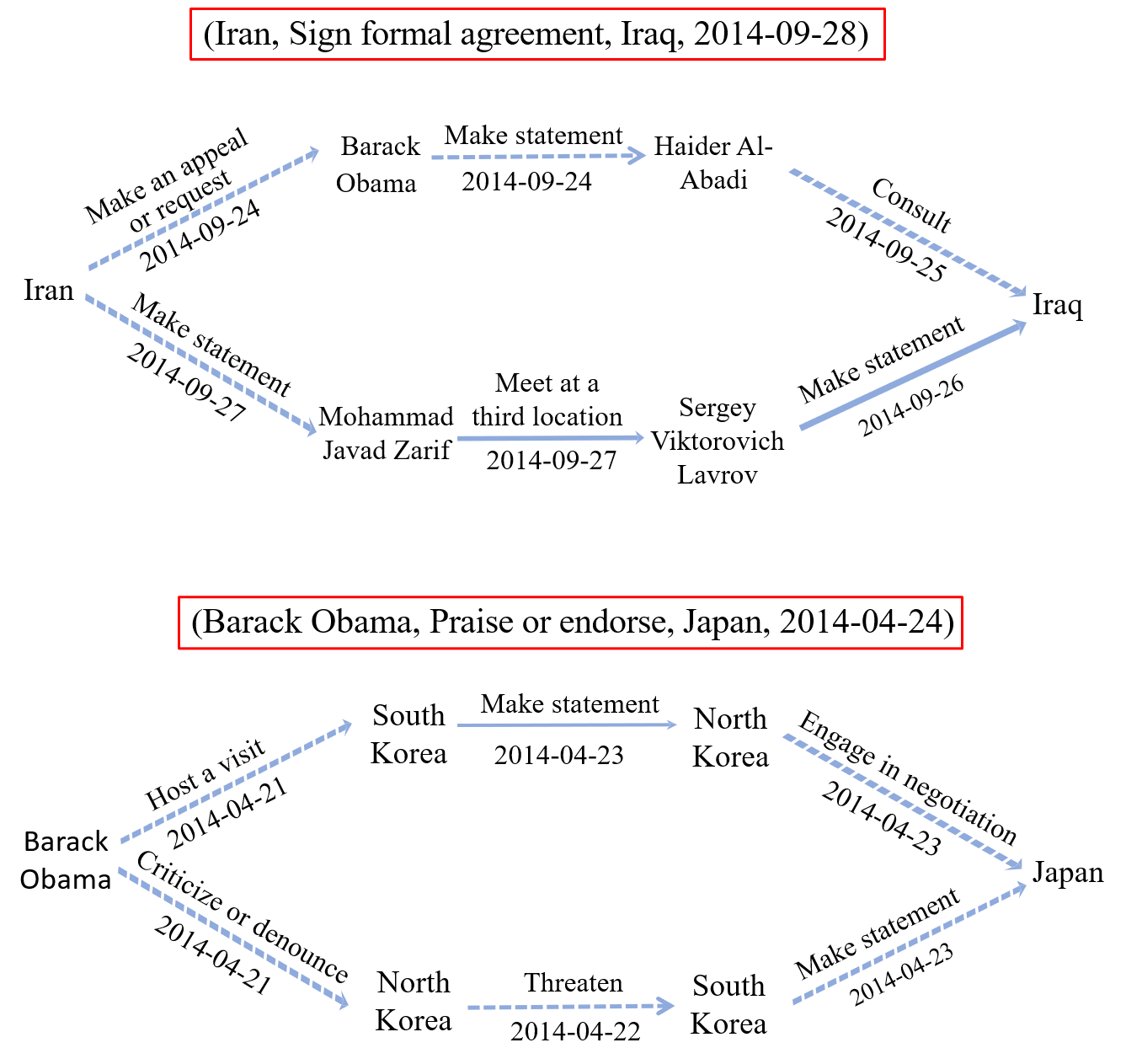}
        \label{fig:case_extra}
    }
    \caption{Path interpretations of TKG reasoning on ICEWS14 test set. Dashed lines mean inverse relations.\label{fig:case}}
\end{figure}

Fig.~\ref{fig:case} illustrates the path interpretations of TKG reasoning for both two settings on the ICEWS14 test set. While users may have different insights towards the visualization, here are our understandings: 1) In Fig.~\ref{fig:case_intro}, for the interpolation setting, when we need to reason about the fact that happened on $t_q =$2014-02-14, we can use the path formed by the links before or after $t_q$. 2) For the extrapolation of Fig.~\ref{fig:case_extra}, we do not strictly impose a chronological order constraint on the links in the path, ensuring that the paths used in our reasoning can be as rich as possible. 3) Multiple different temporal paths may lead to the same reasoning destination, and the explanation is that a reasoning result can have multiple evidence chains.

In addition, we conduct analyses on path lengths, and the contents can be found in Appendix \ref{sec:path_length}.

\subsection{Robustness Analysis}
\begin{figure}[t]
\centering
    \subfigure[Interpolation]{
       \centering
        \includegraphics[width=0.45\linewidth]{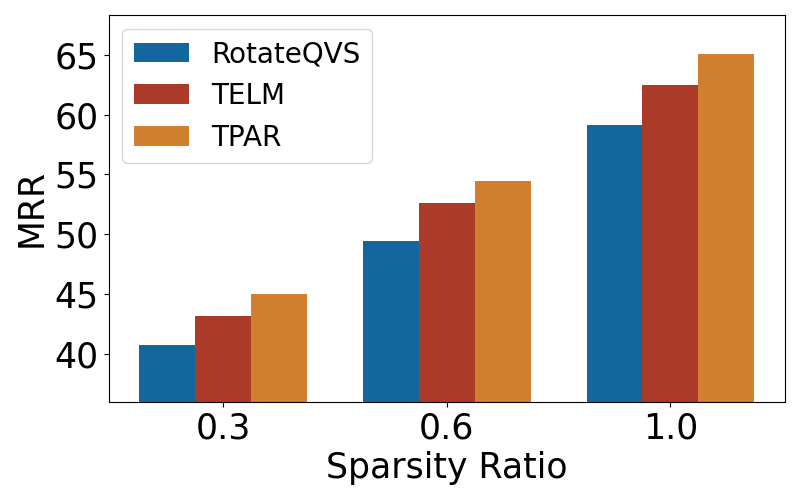}
        \label{fig:sparsity_intro}
    }
    \hspace{-3mm}\subfigure[Extrapolation]{
        \centering
        \includegraphics[width=0.45\linewidth]
        {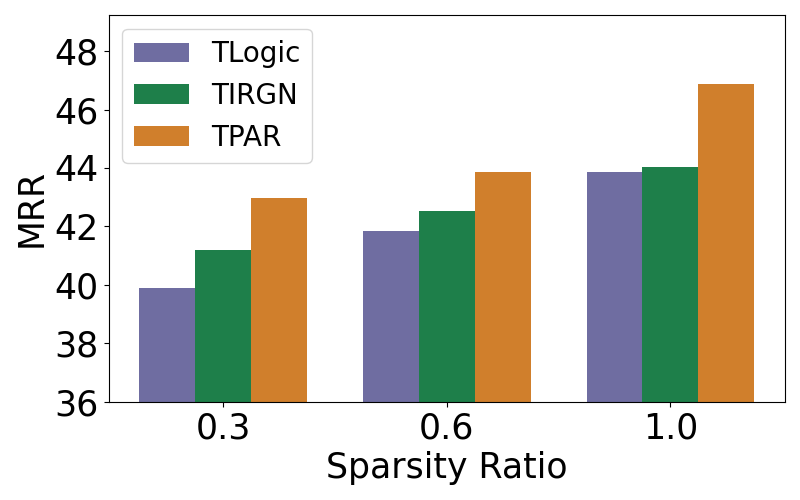}
        \label{fig:sparsity_extra}
    }
    \caption{Performance w.r.t. sparsity ratio on ICEWS14. \label{fig:sparsity}}
\end{figure}
Fig.\ref{fig:sparsity} illustrates a performance comparison under varying degrees of data sparsity on the ICEWS14 dataset. We compare our TPAR model with RotateQVS \cite{ChenWLL22} and TELM \cite{XuCNL21} in the interpolation setting, while TLogic \cite{LiuMHJT22} and TIRGN \cite{LiS022} are taken as extrapolation baselines.
Our TPAR demonstrates superior performance compared to the other models across a wide range of data sparsity levels for both interpolation and extrapolation settings. This highlights its exceptional robustness and ability to handle sparse data effectively.

Furthermore, we conduct an analysis on the chronological order in Appendix \ref{sec:ablation_time_constrain}, which also demonstrates robustness to some extent.

\section{Conclusion}
We propose a temporal path-based reasoning (TPAR) model with a neural-symbolic fashion that can be applied to both interpolation and extrapolation TKG Reasoning settings in this paper. Introducing the Bellman-Ford Algorithm and a recursive encoding method to score the destination entities of various temporal paths, TPAR performs robustly on uncertain temporal knowledge with fine interpretability. Comprehensive experiments demonstrate the superiority of our proposed model for both settings. A well-designed pipeline experiment further verified that TPAR is capable of integrating two settings for a unified TKG reasoning.


\section{Limitations}
We identify that there may be some possible limitations in this study. First, our reasoning results are dependent on the embeddings learned by the proposed (Bellman-Ford-based) recursive encoding algorithm, where the selection of hyperparameters may have a significant impact on the reasoning results.
Further, we can explore additional neural-symbolic reasoning approaches, expanding beyond the current neural-driven symbolic fashion.
Second, the reasoning process relies on attention mechanisms, but it has been pointed out \cite{SuiWWL0C22} that existing attention-based methods are prone to visit the noncausal features as the shortcut to predictions. In future work, we will try to focus on researching causal attention mechanisms for a more solid comprehension of the complicated correlations between temporal knowledge.

\section*{Acknowledgements}
This work is particularly supported by the National Key Research and Development Program of China (No. 2022YFB31040103), and the National Natural Science Foundation of China (No. 62302507).

\bibliography{reference}
\bibliographystyle{acl_natbib}

\clearpage
\appendix

\section{Bellman-Ford Shortest Path Algorithm}\label{sec:Bellman-Ford}
\begin{algorithm}[!ht]
	\caption{The Bellman-Ford Shortest Path Algorithm}
	\label{alg:BF}
	\renewcommand{\algorithmicrequire}{\textbf{Input:}}
	\renewcommand{\algorithmicensure}{\textbf{Output:}}
	\begin{algorithmic}[1]
		\REQUIRE A graph $G$ with $n$ nodes, the edge set $E$, the source node $s$  
		\ENSURE The distance from node s to any node in the graph    
		
		\STATE  $d^{(0)} \gets [+\infty, +\infty, ..., +\infty]    $
		\STATE $d^{(0)}[s] \gets 0    $
		\FOR{$i = 1$ \TO $n-1$}
		\STATE $d^{(i)} \gets d^{(i-1)}    $
		\FOR{each edge $(u, v) \in E$} 
		\STATE \small{$d^{(i)}[v] \gets \min \left\{d^{(i)}[v], d^{(i-1)}[u]+w(u, v)\right\}$ }
		\ENDFOR
		\ENDFOR
		\RETURN $d^{(n-1)}$
	\end{algorithmic}
\end{algorithm}

The Bellman-Ford shortest path algorithm \cite{Ford1956NETWORKFT,Bellman1958ONAR,2010Baras}, founded by Richard Bellman and Lester Ford, is an efficient algorithm for searching the shortest path from the starting point to each node in a graph. The details are shown in Algorithm \ref{alg:BF}. And the basic idea is: For a graph with $n$ nodes, the shortest path between two nodes in the graph can contain at most $n-1$ edges. Then, we can perform $n-1$ times of relaxation operations recursively on the graph to obtain the shortest path from the source node to all nodes in the graph.

\section{An Illustration of Temporal Paths and Temporal Links}
\label{sec:illustraion}

\begin{figure}[ht]
\centering
\includegraphics[width=0.95\columnwidth]{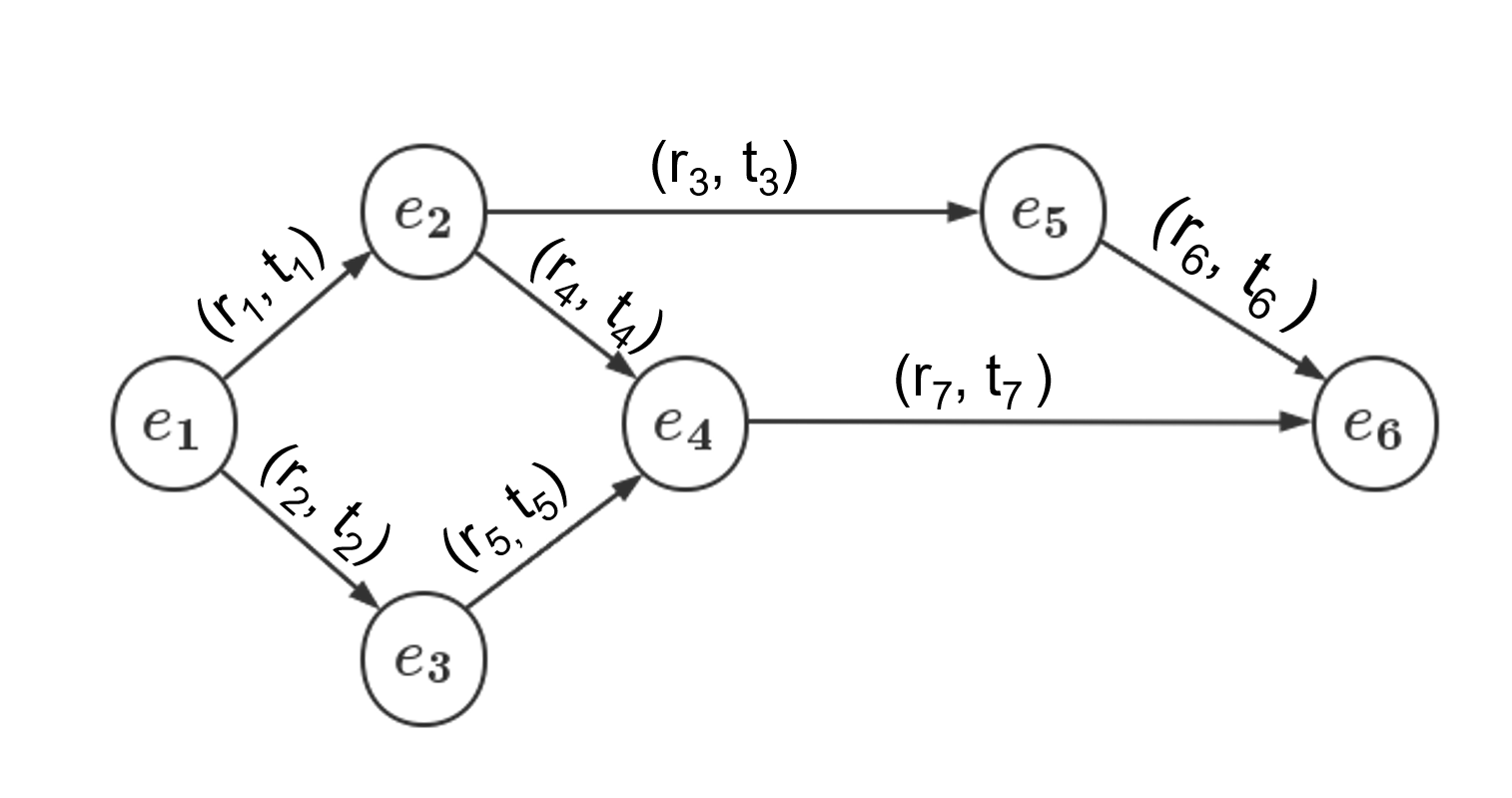}
\caption{An illustration of temporal paths and temporal links.}
\label{figure:path}
\end{figure}

Fig.~\ref{figure:path} illustrates some temporal paths and temporal links. For example, $\small{\begin{matrix} t_1 \\ r_1 \end{matrix} (e_1, e_2)}$ is a temporal link pointing from $e_1$ to $e_2$, while $\small{\begin{matrix} t_2 \\ r_2 \end{matrix} (e_1, e_3) \wedge \begin{matrix} t_5 \\ r_5 \end{matrix} (e_3, e_4) \wedge \begin{matrix} t_7 \\ r_7 \end{matrix} (e_4, e_6)}$ and $\small{\begin{matrix} t_1 \\ r_1 \end{matrix} (e_1, e_2) \wedge \begin{matrix} t_3 \\ r_3 \end{matrix} (e_2, e_5) \wedge \begin{matrix} t_6 \\ r_6 \end{matrix} (e_5, e_6)}$ are two distinct temporal paths from $e_1$ to $e_6$.

\section{Benchmark Datasets}
\label{sec:app_datasets}
\subsection{Interpolation datasets}

\begin{table*}[ht]
\centering
\begin{minipage}[t]{0.98\textwidth}
\centering
\resizebox{0.98\textwidth}{!}{
\begin{tabular}{l|rrrr|rrrr}
\toprule
\multirow{2}{*}{Dataset}  & \multicolumn{4}{c|}{\textbf{{Interpolation}}}   & \multicolumn{4}{c}{\textbf{Extrapolation}}  \\
\cline{2-9}
 & ICEWS14  & YAGO11k& ICEWS05-15 & Wikidata12k & ICEWS14 & ICEWS18 & YAGO & WIKI \\ 
 \hline
{Entities} & 7,128      & 10,623  & 10488 & 12554 & 6,869  & 23,033   &10,623  & 12,554  \\
{Relations}  & 230    & 10  & 251 & 24 & 230 &256  &10 & 24  \\
{Train} & 72,826    & 16,408 & 386,962  & 32,497 & 74,845 & 373,018 & 161,540 & 539,286\\
{Validation}  & 8,941  & 2,050 & 46,275  & 4,062 & 8,514 & 45,995 & 19,523 & 67,538\\
{Test}  & 8,963  & 2,051 & 46,092 & 4,062 & 7,371 & 49,545 & 20,026 & 63,110
\\
{Time granularity}  & 24 hours    & 1 year & 24 hours   & 1 year & 24 hours & 24 hours & 1 year & 1 year \\
\bottomrule
\end{tabular}
}
\end{minipage}
\caption{Statistics of our experimented datasets for both interpolation setting and extrapolation setting.}
\label{table:datasets}
\end{table*}

To evaluate our proposed TKG Reasoning method, we perform link prediction task on four widely-used interpolation datasets, namely {ICEWS14}, {{ICEWS05-15} \cite{garcia2018learning}, {YAGO11k} \cite{dasgupta2018hyte} and {WIKIDATA12k} \cite{dasgupta2018hyte}.\footnote{All of the four interpolation datasets can be downloaded from \url{https://github.com/soledad921/ATISE}.}
The left half of Table \ref{table:datasets} summarizes the details of the four interpolation datasets that we use.

ICEWS \cite{lautenschlager2015icews} is a repository containing political events with a specific timestamp. ICEWS14 and ICEWS05-15 \cite{garcia2018learning} are two subsets of ICEWS corresponding to facts from 1/1/2014 to 12/31/2014 and from 1/1/2005 to 12/31/2015.
YAGO11k and WIKIDATA12k \cite{dasgupta2018hyte} are subsets of YAGO3 \cite{Mahdisoltani2015YAGO3AK} and WIKI \cite{ErxlebenGKMV14}, where time annotations are represented as time intervals.
We derive the two datasets from HyTE \cite{dasgupta2018hyte} to obtain the same year-level granularity by dropping the month and date information. And the characteristic of a dataset used for the interpolation is disorderly arrangement, which means that it is random and not sorted according to time.

\subsection{Extrapolation datasets}
The right half of Table \ref{table:datasets} summarizes the details of the four extrapolation datasets that we use, namely {ICEWS14} \cite{garcia2018learning}, {ICEWS18} \cite{jin2020recurrent}, {YAGO} \cite{Mahdisoltani2015YAGO3AK} and {WIKI} \cite{leblay2018deriving}.\footnote{All of the four extrapolation datasets can be downloaded from \url{ https://github.com/Lee-zix/RE-GCN}.}
{ICEWS14} \cite{garcia2018learning} and {ICEWS18} \cite{jin2020recurrent} are two subsets of ICEWS \cite{lautenschlager2015icews} corresponding to facts from 1/1/2014 to 12/31/2014 and from 1/1/2018 to 10/31/2018.
We drop the month and date information of YAGO and WIKI here for ease of processing and to obtain the same year-level granularity as \cite{jin2020recurrent}.
Compared to the interpolation datasets (see the left half of Table \ref{table:datasets}), the facts in the four datasets for extrapolation are arranged in ascending order according to time.
For the purpose of extrapolation, we follow \citet{HanCMT21} and split each dataset into three subsets by time, ensuring (time of training set) < (time of validation set) < (time of test set). Thus, we can smoothly use past knowledge (facts in the training set) to predict the future.

Some recent works have shown that the time-unwise filtered setting \cite{bordes2013translating} may probably get incorrect higher ranking scores \cite{LiJLGGSWC21}, and that the time-wise filtered setting \cite{xu2019temporal,goel2020diachronic} is more suitable for TKG Reasoning since it ensures the facts which do not appear at time $t$ are still considered as candidates for evaluating the given test quadruple \cite{xu2020tero}. Therefore, we use the time-wise filtered setting \cite{xu2019temporal,goel2020diachronic} to report the experimental results.

\section{Experimental Details for Interpolation Reasoning}
\label{sec:appendix_inter}

\subsection{Setup}\label{sec:intro_setup}
In this setting, we split each dataset into three subsets: a training set $\mathcal{T}_{\text {train}}$, a validation set $\mathcal{T}_{\text {valid}}$ and a test set $\mathcal{T}_{\text {test}}$.

For training, we further split the training set $\mathcal{T}_{\text {train}}$ into two parts: the fact set $\mathcal{F}_{\text {train}}$ including $3/4$ of the training quadruples used to extract temporal paths, and the query set $\mathcal{Q}_{\text {train}}$ including the other $1/4$ of the training quadruples used as queries. Thus, we can train our model by using temporal paths collected by $\mathcal{F}_{\text {train}}$ to answer queries in $\mathcal{Q}_{\text {train}}$. 

For verification and testing, we accomplish reasoning by answering queries derived from the validation set and test set, where the whole training set $\mathcal{T}_{\text {train}}$ is prepared to extract and collect temporal paths. Since the entire training set is known to us and we do not use any data from validation or test set during the training process, we can ensure that there is no data leakage issue.

\subsection{Baselines}
For interpolation TKG Reasoning baselines, we consider TTransE \cite{leblay2018deriving}, HyTE \cite{dasgupta2018hyte}, TA-DistMult \cite{garcia2018learning}, ATiSE \cite{xu2019temporal}, and TeRo \cite{xu2020tero},  TComplEx \cite{LacroixOU20}, T-GAP \cite{JungJK21}, TELM \cite{XuCNL21}, RotateQVS \cite{ChenWLL22} and TGeomE++ \cite{XuNCL23}.

\subsection{Hyperparameter}\label{sec:hyperparameter_intro}
To seek and find proper hyperparameters, we utilize a grid search empirically over the following ranges for all four datasets:
embedding dimension $d$ in $\{32,64,128\}$, learning rate in $[10^{-5},10^{-3}]$, dimension for attention $d_a$ in $[3,4,5]$, dropout in $[0.1,0.2,0.3]$, batch size in $[5,10,20,50]$, the maximum length of temporal paths $L$ in $[3,4,5]$, activation function $\delta$ in [identity, tanh, ReLU], and the optimizer we use is Adam.

And we have found out the best hyperparameters combination as follows: for ICEWS14, we set learning rate as $0.0003$, batch size as $10$, and $\delta$ as $identity$; for ICEWS05-15, we set learning rate as $0.00005$, batch size as $5$, and $\delta$ as $identity$; for YAGO11k, we set learning rate as $0.00005$, batch size as $20$, and $\delta$ as $ReLU$; for WIKIDATA12k, we set learning rate as $0.00002$, batch size as $20$, and $\delta$ as $ReLU$; and for all of four datasets, we choose $L$ as $5$, $d_a$ as $5$, dropout as $0.2$, and $d$ as $128$.

\section{Experimental Details for Extrapolation Reasoning}
\label{sec:appendix_extra}

\subsection{Setup}
In this setting, first of all, we need to arrange each dataset in ascending order according to time.
In addition to the same splitting operation as in Section \ref{sec:intro_setup}, we need to ensure: (time of training set) < (time of validation set) < (time of test set). Since some entities and relations in the validation or test set may not have appeared in the training set, our extrapolation setting can actually be seen as an inductive inference \cite{SunZMH021}.

We use the ground truths for our extrapolation TKG Reasoning, as is the case with many previous methods \cite{jin2020recurrent,LiJLGGSWC21,SunZMH021}. Specifically, for all of the training, validation and testing, we predict future events assuming ground truths of the preceding events are given at inference time \cite{HanCMT21}.

\subsection{Baselines}
For extrapolation TKG Reasoning baselines, we consider RE-NET \cite{jin2020recurrent}, CyGNeT \cite{ZhuCFCZ21}, TANGO \cite{HanDMGT21}, xERTE \cite{HanCMT21}, RE-GCN \cite{LiJLGGSWC21}, TITer \cite{SunZMH021}, CEN \cite{LiGJPL000GC22}, TLogic \cite{LiuMHJT22}, TIRGN \cite{LiS022} and RPC \cite{0006MLLTWZL23}.

\subsection{Hyperparameter}
A grid search has been taken empirically for the extrapolation TKG Reasoning on the same hyperparameter ranges as in Section \ref{sec:hyperparameter_intro}.
And we have found out the best hyperparameters combination as follows: for ICEWS14, we set learning rate as $0.0003$, batch size as $10$, $d$ as $128$, and $\delta$ as $identity$; for ICEWS18, we set learning rate as $0.00005$, batch size as $5$, $d$ as $64$, and $\delta$ as $ReLU$; for YAGO, we set learning rate as $0.00003$, batch size as $10$, $d$ as $64$, and $\delta$ as $identity$; for WIKI, we set learning rate as $0.00003$, batch size as $5$, $d$ as $64$, and $\delta$ as $identity$; and for all of four datasets, we choose $L$ as $5$, $d_a$ as $5$, and dropout as $0.2$.

\section{Details of the Pipeline Setting}\label{sec:detailed_pipeline}
To conduct our experimental pipeline, we sample a certain ratio (60\% for Setting A, 70\% for Setting B, and 80\% for Setting C)  of facts to train interpolation, while the other unsampled facts serve to provide incomplete quadruples (with either the subject entity or object entity masked) to be completed. The completed quadruples are then treated as new knowledge. Finally, the newly acquired knowledge from interpolation could be combined with the interpolation training set, ordered chronologically, and used as the training set for downstream (extrapolation) tasks, with the test set of the standard extrapolation dataset used to evaluate experimental results.

\begin{table}[ht]
 \renewcommand\arraystretch{1.1}
\centering
\begin{minipage}[t]{0.98\columnwidth}
\centering
\resizebox{0.98\columnwidth}{!}{

\begin{tabular}{c|c|c|c|c|c|c}
\toprule
 Ratio & Inter- & Extra- & MRR & Hits@1 & Hits@3 & Hits@10
\\\hline
 60\% & - & TLogic & 41.85 & 32.51 & 47.05 & 59.58
   \\
 60\% & TELM & TLogic & 40.88 & 31.66 & 45.86 & 58.69
   \\
 60\% & - & TIRGN & 42.53 & 32.04 & 48.10 & 62.16 
   \\
 60\% & RotateQVS & TIRGN & 40.86 & 30.24 & 45.78 & 61.63
   \\
 60\% & - & TPAR & 43.86 & 33.81 & 49.04 & 62.83
   \\
 60\% & TPAR & TPAR & 44.56 & 34.68 & 49.68 & 63.32
   \\ \hline 
 70\% & - & TLogic & 41.88 & 32.51 & 46.91 & 59.97
   \\
 70\% & TELM & TLogic & 44.55 & 32.29 & 46.55 & 59.53
   \\
 70\% & - & TIRGN & 43.68 & 33.23 & 48.99 & 63.59 
   \\
 70\% & RotateQVS & TIRGN & 42.35 & 31.87 & 47.78 & 62.25
   \\
 70\% & - & TPAR & 45.49 & 35.69 & 50.33 & 64.30
   \\
 70\% & TPAR & TPAR & 46.07 & 35.92 & 51.06 & 65.02
   \\ \hline 
 80\% & - & TLogic & 42.18 & 32.81 & 47.27 & 60.18
   \\
 80\% & TELM & TLogic & 42.09 & 32.69 & 47.31 & 59.78
   \\
 80\% & - & TIRGN & 43.69 & 32.97 & 49.10 & 64.28
   \\
 80\% & RotateQVS & TIRGN & 42.78 & 32.15 & 48.06 & 63.14
   \\
 80\% & - & TPAR & 46.36 & 36.42 & 51.34 & 65.45
   \\
 80\% & TPAR & TPAR & 46.66 & 36.68 & 51.82 & 65.54
   \\
\bottomrule
\end{tabular}

}
\end{minipage}
\caption{A detailed version of Table \ref{table:pipeline}.}
\label{table:pipeline_full}
\end{table}

\begin{table*}[ht]
 \renewcommand\arraystretch{1.1}
\centering
\begin{minipage}[t]{0.9\textwidth}
\centering
\resizebox{0.9\textwidth}{!}{

\begin{tabular}{l|l}
\toprule
\multirow{3}{*}{\makecell{Step 1 \\(Interpolation)}}
  & \textbf{Inputs:} A ratio (denoted as $r$) of known facts sampled from $\mathcal{G}$, denoted as $r \mathcal{G}$.
\\
 & \makecell[l]{\textbf{Queries:} The other unsampled facts $(1-r) \mathcal{G}$ with either the subject or the \\ \qquad object entity masked.}
 \\
& \textbf{Outputs:} Newly acquired facts $\mathcal{G}_{new}$ by completing the queries.
   \\ \hline
\multirow{3}{*}{\makecell{Step 2 \\(Extrapolation)}}  & \textbf{Inputs:} The basic $r \mathcal{G}$, and the newly acquired $\mathcal{G}_{new}$.
\\
 & \makecell[l]{\textbf{Queries:} The standard extrapolation test set.}
 \\
& \textbf{Outputs:} Performance evaluation.
   \\
\bottomrule
\end{tabular}

}
\end{minipage}
\caption{The process of the proposed pipeline.}
\label{table:process_pipeline}
\end{table*}
The process of the proposed pipeline can be seen in Table \ref{table:process_pipeline}.

The detailed results of the pipeline setting can be found in Table \ref{table:pipeline_full}, where a concise version has been shown in Table \ref{table:pipeline} in Section \ref{sec:pipeline}.

\section{Analysis on Path Length}
\label{sec:path_length}

\begin{figure}[ht]
\centering
\includegraphics[width=\columnwidth]{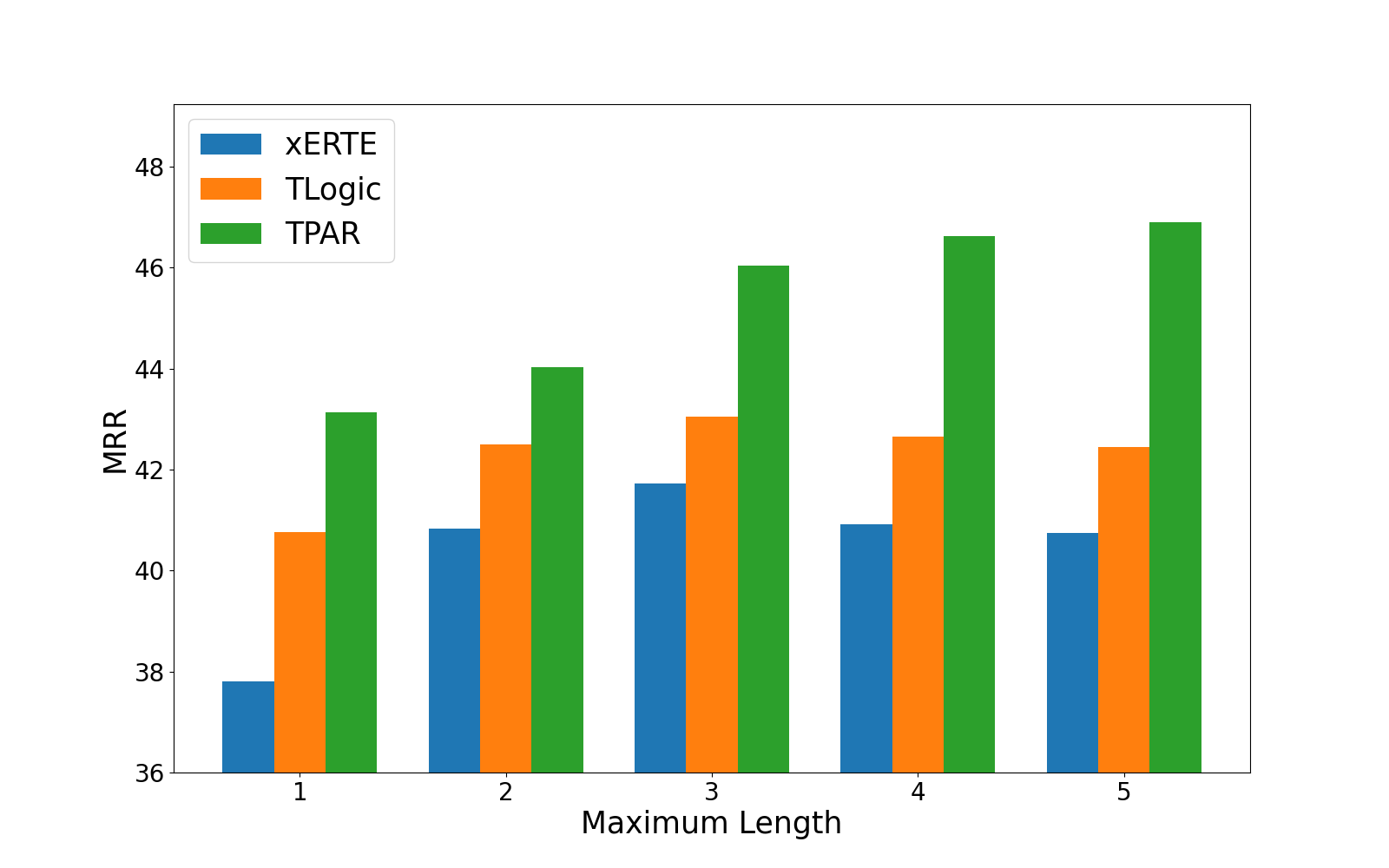}
\caption{The MRR performance with different maximum path lengths for the extrapolation reasoning on the ICEWS14 test set.}
\label{figure:length}
\end{figure}

Fig.~\ref{figure:length} displays the MRR performance with different maximum path lengths for the extrapolation reasoning on ICEWS14 test set. We compare our TPAR with the symbolic method TLogic \cite{LiuMHJT22}, where the maximum length $L$ means it contains all lengths not exceeding $L$ ($L=1,2,\cdots 5$).

As the maximum length $L$ increases, a greater number of temporal links are covered and can provide more detailed information. However, this also makes it more challenging to learn and reason with this information. The results demonstrate that the performance of TLogic decreases for $L$ values of 4 or higher. In contrast, when L is too small, such as $L \leq 2$, our TPAR approach struggles to perform well due to the limited amount of information encoded in these short temporal paths. An interesting discovery can be made: For symbolic TLogic, a balance needs to be struck between the richness of the information contained and the ability to effectively learn from it. However, our TPAR appears to alleviate this issue, as neural-symbolic reasoning is more tolerant of ambiguous and noisy data.

\begin{table}[ht]
\centering
\begin{minipage}[t]{0.9\columnwidth}
\centering
\resizebox{0.9\columnwidth}{!}{

\begin{tabular}{c|cccc}
\toprule
Length & Hits@1  & Hits@3 & Hits@10 & MRR
\\ \hline
1 & 10.03 & 10.45 & 10.58 & 10.32
\\ \hline
2& 10.39 & 11.06 & 11.45 & 10.84
\\ \hline
3 & 15.13 & 22.03 & 32.16 & 21.03
\\ \hline
4& 15.59 & 22.56 & 33.15 & 21.62
\\ \hline
5 & 17.35 & 25.14 & 37.42 & 24.12
    \\
\bottomrule
\end{tabular}

}
\end{minipage}
\caption{Performance of our proposed TPAR with different maximum path lengths for the interpolation reasoning on the YAGO11k test set.}
\label{table:length_inter}
\end{table}

We also take a similar study on the interpolation reasoning, and Table \ref{table:length_inter} illustrates the performance of our proposed TPAR with different maximum path lengths for the interpolation reasoning on the YAGO11k test set. As the length increases, all metrics of our TPAR have significantly improved.

\section{Analysis on Chronological Order}\label{sec:ablation_time_constrain}
\begin{table}[ht]
 \renewcommand\arraystretch{1.1}
\centering
\begin{minipage}[t]{0.95\columnwidth}
\centering
\resizebox{0.95\columnwidth}{!}{

\begin{tabular}{c|c|c|c|c|c}
\toprule
\multicolumn{2}{c|}{Model} & \multicolumn{2}{c|}{\bfseries{TLogic}} &\multicolumn{2}{c}{\bfseries{TPAR}} \\\hline
\multicolumn{2}{c|}{Chronological Order} & \XSolid  & \Checkmark& \XSolid & \Checkmark \\\hline
\multirow{4}{*}{ICEWS14} & Hits@1 & 32.60 & 33.56 & 36.88 & 36.02
    \\\cline{2-6}
 & Hits@3 & 47.06 & 48.27 & 52.28 & 51.49
    \\\cline{2-6}
 & Hits@10 & 60.06 & 61.23 & 65.89     & 64.30
    \\\cline{2-6}
 & MRR & 42.02 & 43.04 & 46.89  & 45.77   
    \\\hline
 \multirow{4}{*}{ICEWS18}  & Hits@1  & 19.92 & 20.54 & 26.58 & 25.24
 \\\cline{2-6}
 & Hits@3 & 33.04 & 33.95 & 39.27 & 38.55
 \\ \cline{2-6}
  & Hits@10 & 47.75 & 48.53 & 56.94 & 55.54
  \\\cline{2-6}
   & MRR  & 28.79 & 29.82 & 35.76 & 33.98
   \\
\bottomrule
\end{tabular}

}
\end{minipage}
\caption{An analysis on chronological order for the extrapolation reasoning.}
\label{table:time_constrains}
\end{table}

As discussed in Section \ref{section:path}, for extrapolation reasoning, our TPAR does not require strict chronological order between links in a path. Alternatively, methods like TLogic \cite{LiuMHJT22} demand that all links in a path be ordered chronologically. For example, given a path $\mathcal{P}= \small{\begin{matrix} t_1 \\ r_1 \end{matrix} (e_1, e_2)} \wedge \small{\begin{matrix} t_2 \\ r_2 \end{matrix} (e_2, e_3)} \wedge \small{\begin{matrix} t_3 \\ r_3 \end{matrix} (e_3, e_4)} $, TLogic requires $t_q>t_1 \geq t_2 \geq t_3$, while we simply require $t_q>t_i$ for $i=1,2, 3$. We conduct an analysis on the chronological order of links in a path, and the results are shown in Table \ref{table:time_constrains}. 
The results indicate that when chronological order is relaxed, the performance of TLogic decreases, while that of our TPAR increases. We believe that by relaxing chronological order in paths, we can gather more paths, thereby providing more information for inference. Nonetheless, this may introduce noise and uncertainty into the process, which ultimately reduces the effectiveness of symbolic methods such as TLogic.
On the other hand, our TPAR employs a neural-symbolic approach, which not only mitigates the impact of these challenges but also ensures excellent inference performance, showcasing its strong robustness.

\end{document}